%% file: acl_latex.tex
\pdfoutput=1
\documentclass[11pt]{article}

\usepackage[preprint]{acl}

\usepackage{times}
\usepackage{latexsym}

\usepackage{multicol}
\usepackage{multirow}
\usepackage{amsmath}
\usepackage{subcaption}
\newcommand{\hytt}[1]{\texttt{\hyphenchar \font=\defaulthyphenchar #1}}
\usepackage[ruled,vlined]{algorithm2e}
\usepackage[T1]{fontenc}
\usepackage{wasysym}
\usepackage{times}
\usepackage{latexsym}
\usepackage[many]{tcolorbox}
\usepackage[T1]{fontenc}
\usepackage{inconsolata}
\usepackage{graphicx}
\usepackage{microtype}
\usepackage{amssymb}
\usepackage{mathtools}
\usepackage{amsthm}
\usepackage{makecell}
\usepackage{booktabs}
\usepackage{array}
\usepackage{longtable}
\usepackage{subcaption}
\usepackage{fontawesome}
\usepackage{centernot}
\usepackage{hyperref}

\usepackage[utf8]{inputenc}

\usepackage{microtype}

\usepackage{inconsolata}

\usepackage{graphicx}
\usepackage{paralist, tabularx}

\definecolor{green}{RGB}{0,150,10}
\definecolor{blue}{RGB}{0,148,181}
\definecolor{orange}{RGB}{194,153,107}

\newcommand{\github}{\raisebox{-1.5pt}{\includegraphics[height=1em]{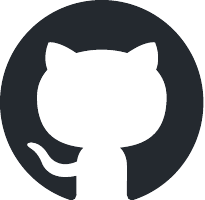}}}
\newcommand{\huggingface}{\raisebox{-1.5pt}{\includegraphics[height=1em]{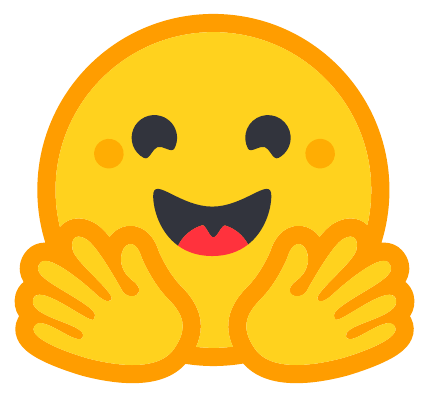}}}

\title{\textsc{SurveyForge}: On the Outline Heuristics, Memory-Driven Generation, and Multi-dimensional Evaluation for Automated Survey Writing}

\author{%
  \textbf{Xiangchao Yan}\textsuperscript{$\clubsuit$,}\footnotemark[1]\;\;%
  \textbf{Shiyang Feng}\textsuperscript{$\clubsuit$,$\spadesuit$,}\footnotemark[1]\;\;%
  \textbf{Jiakang Yuan}\textsuperscript{$\clubsuit$,$\spadesuit$}\;\;
  \textbf{Renqiu Xia}\textsuperscript{$\clubsuit$,$\diamondsuit$}\\[1ex]%
  \textbf{Bin Wang}\textsuperscript{$\spadesuit$}\;\;%
  \textbf{Bo Zhang}\textsuperscript{$\clubsuit$,}\footnotemark[2]\;\;%
  \textbf{Lei Bai}\textsuperscript{$\clubsuit$,}\footnotemark[2]\\[1ex]%
  \textsuperscript{$\clubsuit$}Shanghai Artificial Intelligence Laboratory \; 
  \textsuperscript{$\spadesuit$}Fudan University \; 
  \textsuperscript{$\diamondsuit$} Shanghai Jiao Tong University\\[1ex]%
  \normalsize{\texttt{\{zhangbo,bailei\}@pjlab.org.cn}}\\[1ex]%
  {\github\ \texttt{\url{https://github.com/Alpha-Innovator/SurveyForge}}} \\
  {\huggingface\ \texttt{\url{https://huggingface.co/datasets/U4R/SurveyBench}}}
}

\begin{document}
\maketitle
\footnotetext[1]{Core Contributor}
\footnotetext[2]{Corresponding Authors}

\input{latex/section/abstract}
\input{latex/section/intro}
\input{latex/section/related_work}
\input{latex/section/method}

\input{latex/section/exp}

\input{latex/section/conclusion}
\input{latex/section/Limitations}

\section*{Ethics Statement}
This work focuses on the development of an automated framework for survey generation, aiming to assist researchers in efficiently summarizing existing literature. The proposed method relies on publicly available datasets and research papers, ensuring compliance with copyright and intellectual property laws. While the framework is designed to augment human expertise, we encourage users to critically evaluate the generated outputs to ensure their alignment with ethical research practices and to mitigate any potential limitations, such as biases or incomplete summaries.

\section*{Acknowledgement}
The research was supported by Shanghai Artificial Intelligence Laboratory, the Shanghai Municipal Science and Technology Major Project, and Shanghai Rising Star Program (Grant No. 23QD1401000).

\bibliography{custom}
\input{latex/section/supp}

\end{document}

%% file: latex/section/abstract.tex
\begin{abstract}

%In the era of rapid growth of scientific research, survey papers have become an essential resource for researchers, providing an efficient way to understand the developments and trends within a specific domain. With the advancement of large language models (LLMs), researchers have turned to LLM-based tools to automate the generation of summaries and surveys. Despite improving the efficiency of literature research, existing works still suffer from incomplete outlines, low-quality citations, and relatively subjective evaluation. 
% , which is mainly composed of a heuristic outline generation and a memory-driven content generation. 

Survey paper plays a crucial role in scientific research, especially given the rapid growth of research publications. Recently, researchers have begun using LLMs to automate survey generation for better efficiency. However, the quality gap between LLM-generated surveys and those written by human remains significant, particularly in terms of outline quality and citation accuracy.
To close these gaps, we introduce \textsc{SurveyForge}, which first generates the outline by analyzing the logical structure of human-written outlines and referring to the retrieved domain-related articles.~Subsequently, leveraging high-quality papers retrieved from memory by our scholar navigation agent, \textsc{SurveyForge} can automatically generate and refine the content of the generated article. Moreover, to achieve a comprehensive evaluation, we construct SurveyBench, which includes 100 human-written survey papers for win-rate comparison and assesses AI-generated survey papers across three dimensions: reference, outline, and content quality.
Experiments demonstrate that \textsc{SurveyForge} can outperform previous works such as AutoSurvey. 
% We present several examples of auto-generated survey papers produced by \textsc{SurveyForge} at~\textcolor{teal}{\url{https://anonymous.4open.science/r/survey_example-7C37/}}.
\end{abstract}

%% file: latex/section/intro.tex
\section{Introduction}

\begin{figure}[t!]
   \includegraphics[width=\columnwidth]{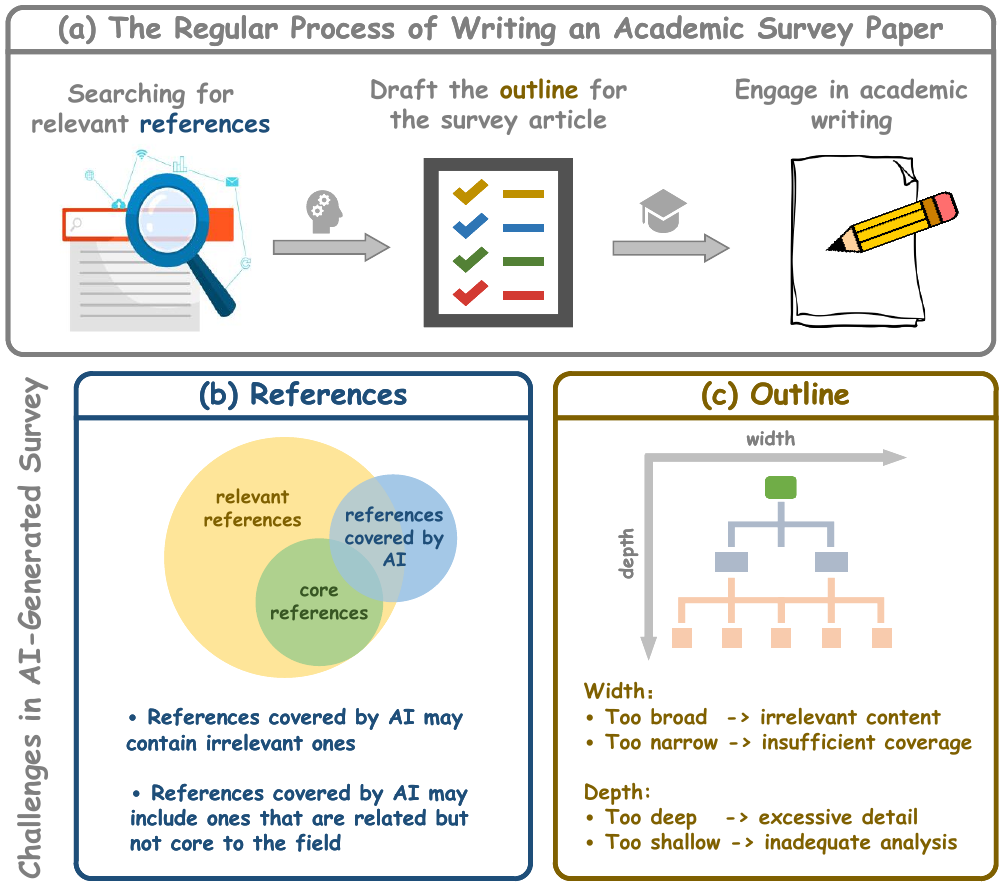}
   \vspace{-0.6cm}
   \caption{Compared to human-written surveys, AI-generated surveys face two primary challenges. First, regarding the outline, these papers may often lack coherent logic and well-structured organization. Second, with respect to references, they frequently fail to include truly relevant and influential literature.}
   \label{fig:intro}
\vspace{-0.5cm}
\end{figure}

With the rapid development of science and technology, the number of published research articles has been growing exponentially, particularly in fast-evolving fields like Artificial Intelligence (AI). The rapid growth of the literature makes it increasingly difficult for researchers to gain in-depth knowledge of a specific scientific field. Survey papers, which systematically integrate existing studies and provide comprehensive developments and trends in the specific domain, have become a vital starting point of the scientific research cycle. However, traditional human-driven survey writing requires researchers to review a vast number of articles which is time-consuming and makes it challenging to keep up-to-date with the latest advancements in the field.

Inspired by the remarkable advancement and capabilities of Large Language Models (LLMs) \citep{achiam2023gpt4,Anthropic2024claude3,touvron2023llama, cai2024internlm2}, researchers have begun utilizing them to automatically review the literature and generate survey papers. As a pioneer, GPT-Researcher \citep{assafelovic2023gptresearcher} generates survey papers based on the abstract of topic-relevant articles retrieved from multiple online academic databases. To identify more relevant literature to the survey topic, AutoSurvey \citep{wang2024autosurvey} constructs a local literature database based on arXiv, establishes vector indices for each literature, and concurrently generates content for each subsection. To further align the writing style of LLM-generated content with that of humans, OpenScholar \citep{asai2024openscholar} proposes a large-scale scientific literature dataset, and fine-tunes the LLMs based on this dataset to obtain a model specifically designed for answering scientific questions.  

Most of these automated survey generation methods follow the traditional academic survey writing workflow: from literature search, to outline drafting, and finally academic writing, as illustrated in Fig.~\ref{fig:intro}. However, despite the promising achievements of the aforementioned methods, several significant challenges still remain. \textbf{Firstly}, the structure of AI-generated surveys often lacks coherent logic and is often poorly-organized. For example, as shown in Fig.~\ref{fig:intro}, existing works may suffer from structural imbalance in both width and depth, such as overly detailed sectioning or inadequate coverage of key topics. %This makes it difficult to capture the logical connections between different sections.
\textbf{Secondly}, AI-generated surveys often fail to reference key influential literature, reducing the overall depth and value of surveys. As shown in Fig.~\ref{fig:intro}, they may cite irrelevant works while overlooking important contributions in the field. \textbf{Lastly}, the evaluation of AI-generated surveys mainly relies on LLMs, focusing on the overall quality of the long-form content. This approach lacks fine-grained analysis of critical aspects such as outline quality, reference relevance, and structural coherence. Moreover, the absence of objective evaluation criteria makes it difficult to establish consistent quality benchmarks or compare different methods effectively.

%generate a summary report, dividing the survey into overly fragmented subsections and treating a single paper as an independent subsection, rather than synthesizing and integrating related studies under a broader academic framework.
%This issue arises primarily because the LLMs lack the prior knowledge and understanding of academic writing conventions that human authors possess. 
% Firstly, from the perspective of the structure of the generated survey paper,... Secondly, in terms of the reference quality,... Besides, The evaluation for generated survey papers previously relied predominantly on human or LLMs' scoring, lacking objective and quantifiable evaluation metrics

To address the aforementioned challenges, we propose an automated framework for generating survey papers, namely \textsc{SurveyForge} which contains two stages: Outline Generation and Content Generation. In the first stage, \textsc{SurveyForge} employs a heuristic learning approach to leverage topic-relevant literature and structural patterns from human-written surveys, generating semantically comprehensive and well-organized outlines. In the second stage, a memory-driven scholar navigation agent, with a temporal-aware reranking engine, retrieves high-quality literature for each subsection. Then, the content for each section is combined and refined into a coherent and comprehensive survey. Furthermore, we construct \textbf{SurveyBench}, a multi-dimensional benchmark to facilitate systematic assessment of automated survey generation systems.

Extensive results highlight the unique strengths of \textsc{SurveyForge} across multiple dimensions, including its ability to generate well-structured outlines, retrieve high-quality and highly relevant references, and produce coherent, comprehensive content. \textsc{SurveyForge} not only delivers measurable improvements in these areas but also demonstrates a remarkable ability to bridge the gap between AI-generated and human-written surveys. These findings underscore its potential as a robust framework for automated survey generation, setting a new standard for quality and reliability in this domain.

Our contribution can be summarized as follows.
\begin{compactitem}
    \item We propose \textsc{SurveyForge}, a novel automated framework for generating high-quality academic survey papers.
    \item We propose a heuristic outline generation method and a memory-driven scholar navigation agent, which together ensure a well-structured survey framework and high-quality content generation.
    \item To facilitate objective evaluation, we establish SurveyBench, a comprehensive benchmark featuring quantifiable metrics for assessing outline quality, reference quality, and content quality.
\end{compactitem}

% The remainder of this paper is organized as follows. A comprehensive overview of the related works is given in Section ?. In Section ?, the proposed method is presented in detail. Section ? presents the experimental results and conducts insightful analysis of our proposed methods. Concluding remarks are presented in Section ?.

\begin{figure*}[t!]
\vspace{-0.6cm}
  \centering
   \includegraphics[width=0.99\linewidth]{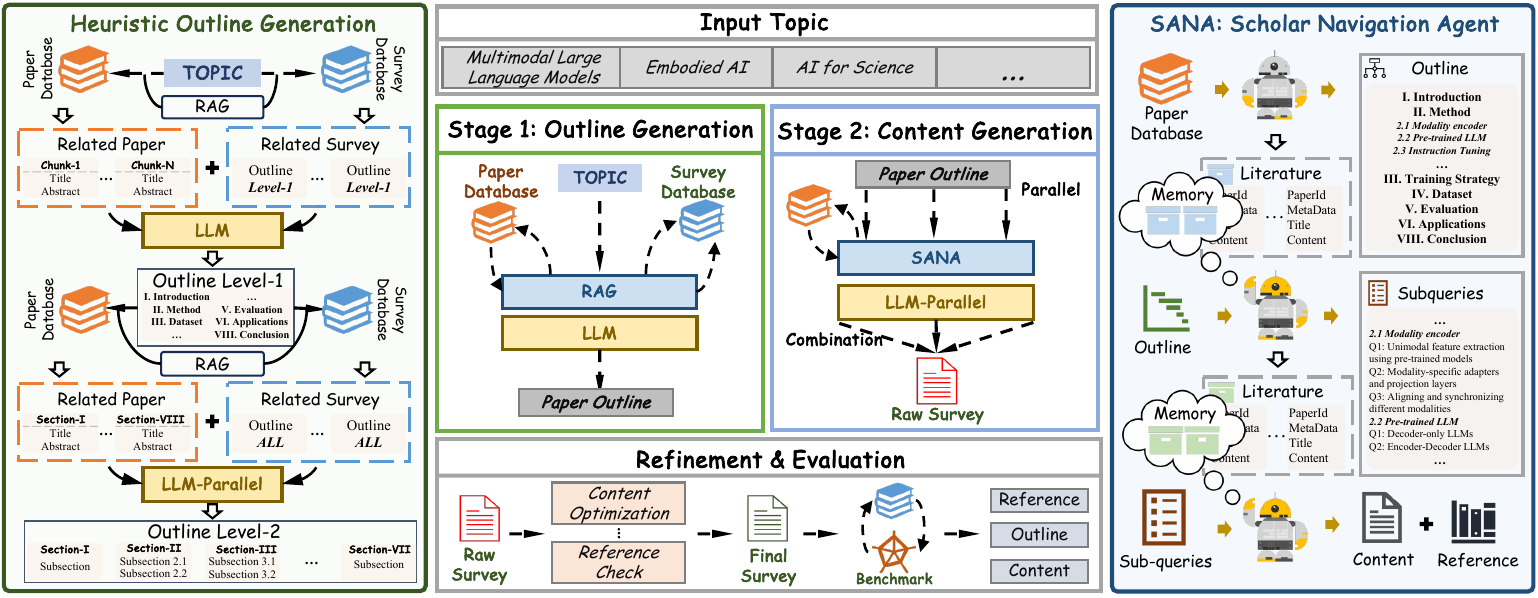}
   \vspace{-0.2cm}
   \caption{The overview of \textsc{SurveyForge}. The framework consists of two main stages: Outline Generation and Content Writing. In the Outline Generation stage, \textsc{SurveyForge} utilizes heuristic learning to generate well-structured outlines by leveraging topic-relevant literature and structural patterns from existing surveys. In the Content Writing stage, a memory-driven Scholar Navigation Agent (SANA) retrieves high-quality literature for each subsection and LLM generates the content of each subsection. Finally, the content is synthesized and refined into a coherent and comprehensive survey.}
   \label{fig:framework}
\vspace{-0.3cm}
\end{figure*}

%% file: latex/section/related_work.tex
\section{Related Work}

\noindent \textbf{Autonomous Scientific Discovery.} With the advancement of LLMs \citep{achiam2023gpt4, Anthropic2024claude3, chen2024far}, an increasing number of researchers have begun exploring their potential for autonomous scientific discovery~\citep{xia2024chartx, li2024chain, xia2024docgenome, huang2024mlagentbench,ghafarollahi2024sciagents, chen2024scienceagentbench}. Several studies~\citep{li2024chain, hu2024nova, kumar2024can, wang2024scipip, su2024two} have focused on leveraging LLMs for novel scientific idea generation. For instance, COI-Agent \citep{li2024chain} introduces an innovative chain-structured literature organization framework. SCIPIP \citep{wang2024scipip} proposes a hybrid approach combining literature-based and brainstorming-based generation to improve both the novelty and feasibility of the generated ideas. Beyond these specific applications, researchers have also developed comprehensive systems for scientific discovery. AI-Scientist~\citep{AIScientist} designs a comprehensive pipeline that covers idea generation, experimental design, and manuscript writing. More recently, Dolphin~\citep{yuan2025dolphin} develops a closed-loop LLM-driven framework to boost the automation level of scientific research.

\noindent \textbf{Automated Survey Generation.} With the rapid proliferation of scientific papers, it has become increasingly challenging for researchers to track developments in specific fields. Early methods \citep{hoang2010towards, hu2014automatic, jha2015content, chen2019automatic} primarily rely on content models to select and organize sentences from papers, often resulting in outputs lacking coherence and readability. Sun et al. \citep{sun2019automatic} introduce a template tree that generates content recursively based on nodes, which improves coherence but remains inflexible. Recognizing the need for more flexible and coherent solutions, the emergence of LLMs has introduced new opportunities for enhancing the automated survey generation. Researchers have begun to leverage LLMs to facilitate efficient literature comprehension and review \citep{wang2024autosurvey, hu2024hireview}. Zhu et al. \citep{zhu2023hierarchical} introduce a novel task of hierarchical catalogue generation for surveys, along with corresponding semantic and structural metrics for evaluation, but it is limited to outline generation with fixed reference papers. AutoSurvey \citep{wang2024autosurvey} proposes a two-stage LLM-based method for survey generation but fails to focus on the analysis of human academic writing styles and key references, which are crucial for producing high-quality surveys. Subsequently, HiReview \citep{hu2024hireview} introduces a taxonomy-driven framework to explore paper relationships hierarchically, enhancing LLMs' understanding of inter-paper connections. However, relying on 2-hop citation networks from existing surveys instead of commonly-cited papers limits its broader applicability.

%% file: latex/section/method.tex
\section{Method}
\vspace{-0.15cm}

In this section, we propose \textsc{SurveyForge}, a novel framework based on LLMs for automatically retrieving relevant literature and generating comprehensive survey papers. As shown in Fig.~\ref{fig:framework}, our framework consists of two main stages: outline generation stage and content writing stage. The outline generation stage leverages both research papers and existing survey structures through a heuristic learning mechanism, producing academically structured outlines. The content generation stage employs a memory-driven scholar navigation agent with key paper retrieval strategy to synthesize the content of the survey. Finally, we propose a benchmark \textbf{SurveyBench} for automated survey generation tasks. The details are elaborated in Sec.~\ref{sec:outline_generation}, Sec.~\ref{sec:content_writing} and Sec.~\ref{sec:benchmark}, respectively.

\subsection{Heuristic Outline Generation}
\label{sec:outline_generation}

% Few-shot Heuristic Learning
The outline of a survey paper is crucial as it defines the logical organization and knowledge structure of the entire work. While LLMs excel at generating textual content, they often fall short in crafting well-structured survey outlines. Common issues include a lack of hierarchical depth, insufficient theoretical grounding, and a tendency toward report-like structures rather than scholarly frameworks. These limitations can be attributed to the limited understanding of academic writing conventions and the organizational principles underlying survey design. To address these challenges, we propose a top-down heuristic learning approach, enabling LLMs to understand the established theoretical frameworks and organizational paradigms from human-written survey outlines. Our approach is underpinned by two domain-specific knowledge bases: a Research Paper Database, which encodes domain knowledge, and a Survey Outline Database, which captures established structural patterns (details provided in Appendix.~\ref{details_databse}). 
As shown in Algorithm~\ref{alg:algorithm}, the framework begins with cross-database knowledge fusion, retrieving relevant papers and outlines for the given topic $T$ from $\mathcal{D}_R$ and $\mathcal{D}_S$. This process identifies key thematic areas and their interrelations, generating the first-level outline $\mathcal{O}_i$ augmented with semantic queries ${Q_i}$ that specify the scope and focus of each heading. For each section $\mathcal{O}_i$, we recursively retrieves relevant materials ($\mathcal{P}_{R_i}$, $\mathcal{P}_{S_i}$) and generates second-level outlines $\mathcal{O}_{ij}$ with sub-queries ${q_{ij}}$. Finally, these headings and their associated queries are systematically merged to construct a academically rigorous and comprehensive survey outline, serving as a foundation for subsequent content generation.

\subsection{Memory-Driven Content Generation}
\label{sec:content_writing}

The memory-driven content generation stage consists of two primary steps: literature retrieval and parallel content creation.~These steps are performed sequentially by the proposed Scholar NAvigation Agent (SANA) and the LLM, respectively. A detailed explanation of each step is provided below.

\subsubsection{SANA: Scholar Navigation Agent}

\begin{algorithm}[t!]
\small
\caption{\textsc{SurveyForge}}
\label{alg:algorithm}
\KwIn{Survey Topic $T$; Research Paper Database $\mathcal{D}_R$; Survey Outline Database $\mathcal{D}_S$}
\KwOut{Final Survey Document $F$}
\BlankLine

\emph{/* Outline Generation */}\\
Retrieve relevant papers and outlines for $T$: $\mathcal{P}_R$, $\mathcal{P}_S$\;
Generate first-level outline $\mathcal{O}_i$ and queries $\{Q_i\}$\;
\ForEach{first-level $\mathcal{O}_i$ }{
    Retrieve relevant papers and outlines for $Q_i$: $\mathcal{P}_{R_i}$, $\mathcal{P}_{S_i}$\;
    Generate second-level outline $\mathcal{O}_{ij}$ and queries $\{q_{ij}\}$\;
    Store $\mathcal{P}_{R_i}$ as memory $M_i$\;
}
Store $\mathcal{P}_R$ as overall memory $M$\;

\BlankLine
\emph{/* Content Generation */}\\
\ForEach{subsection ${O}_{ij}$ \textbf{in parallel}}{
    Decompose query $q_{ij}$ into sub-queries $\{q_{ijk}\}$ using $M_i$\;
    Initialize $L_{ij} \leftarrow \emptyset$\;
    \ForEach{sub-query $q_{ijk}$}{
        Retrieve papers $L_{ijk}$ using $q_{ijk}$ and $M$\;
        $L_{ij} \leftarrow L_{ij} \cup L_{ijk}$\;
    }
    Rerank and select top papers $L_{ij}^{\text{reranked}}$\;
    Generate content $C_{ij}$ for $O_{ij}$ using $L_{ij}^{\text{reranked}}$\;
}

\BlankLine
Merge contents $\{C_{ij}\}$ to form draft $F_{\text{draft}}$\;
Refine $F_{\text{draft}}$ to produce final document $F$\;

\BlankLine
\Return{$F$}\;
\end{algorithm}

To ensure that the quality and quantity of references in the generated survey papers, we propose a Scholar Navigation Agent (SANA), equipped with \textit{memory} and \textit{reranking} capabilities, designed to facilitate literature retrieval across various generation stages. The SANA includes three modules: Memory for Sub-query (MS), Memory for Retrieval (MR), Temporal-aware Reranking Engine (TRE). 

\noindent \textbf{Memory for Sub-query.} 
Query decomposition is a common technique that involves breaking down a complex query into smaller sub-queries, thereby enabling more precise information retrieval. Existing query decomposition methods \citep{fan2024survey} are mostly achieved through naive prompts and LLMs. However, such methods require meticulous tuning of prompts to accommodate different tasks and may cause significant semantic differences between the decomposed sub-queries and the original query, which could potentially degrade the quality of the references in the AI-generated surveys. Therefore, we incorporate the memory mechanism into the query decomposition process of SANA to enhance the effectiveness of sub-queries. 
Specifically, as described in Sec.~\ref{sec:outline_generation}, when generating the first-level outline $O_i$, a set of literature $\mathcal{P}_{R_i}$ is retrieved by Retrieval-Augmented Generation (RAG). In the MS module, SANA takes the literature $\mathcal{P}_{R_i}$ as memory $M_i$, the original query consists of the titles $t_{O_{ij}}$ and descriptions $d_{O_{ij}}$ of each subsection:
\begin{equation}
    q_{ij}=[d_{O_{ij}},t_{O_{ij}}].
\end{equation}
To achieve query decomposition, $q_{ij}$ and $M_i$ are used together as part of the instruction to prompt the LLM to decompose $q_{ij}$ into multiple sub-queries $q_{ijk}$:
\begin{equation}
    q_{ijk} = \mathrm{LLM}(q_{ij}, M_i).
\end{equation}
Finally, the sub-query $q_{ijk}$ is used in the subsequent MR module to retrieve literature related to the sub-section $O_{ij}$.
% Specifically, when generating the first-level outline $O_i$, SANA retrieves a set of references $P_{R_i}$ based on the title $t_{O_i}$ and description $d_{O_i}$ of each section. It then uses these references to generate the titles $t_{O_{ij}}$ and descriptions $d_{O_{ij}}$ for each subsection under that section.

\noindent \textbf{Memory for Retrieval.} The effectiveness of content generation heavily depends on the quality of retrieved information. Traditional retrieval methods \citep{lewis2020retrieval,gao2023retrieval}, which typically query the entire literature database $\mathcal{D}_R$, are often inefficient and lack contextual focus, particularly in generating complex, multi-section documents. These methods treat each section as an isolated unit, failing to account for the global structure and thematic coherence of the document. This results in redundant or irrelevant retrievals and limits the overall coherence of generated content.

To address these limitations, we incorporate the memory mechanism into the retrieval process of SANA to bridge the gap between the outline and content generation stages. Specifically, in the MS module, SANA takes the literature $\mathcal{P}_R$ related to the entire outline as memory $M$. Based on the embedding similarity between each sub-query $q_{ijk}$ and the literature in $M$, the most relevant literature $L_{ijk}$ for each sub-query of section $O_{ij}$ is retrieved. Subsequently, the retrieved literature $L_{ijk}$ is reranked and selected within the following TRE module for content generation.

\noindent \textbf{Temporal-aware Reranking Engine.} 
Reranking plays a important role in enhancing the quality and relevance of retrieved information. Existing methods \citep{glass-etal-2022-re2g,bge_embedding} typically employ advanced scoring mechanisms to measure textual relevance between queries and documents. However, these surface-level semantic matching may fall short in capturing the academic impact and quality of publications. Besides, The publication date of a paper plays a critical role in determining its influence and significance within its respective field. Consequently, analyzing papers from different time periods within the same research domain is a crucial for identifying high-quality contributions in the research field. For papers published within the same time period, there are various metrics to indicate their impact and quality, such as citation count, Essential Science Indicators (ESI), etc \citep{esi}. Among these, citation count serves as a complementary quality indicator that reflects the scholarly influence and recognition of research works. To address both the limitations of pure semantic matching and the temporal bias in different quality indicators, we propose a temporal-aware reranking engine that integrates textual relevance, citation impact, and publication recency. This approach ensures not only the topical relevance but also the academic quality of the retrieved literature while maintaining a balanced representation of both established and emerging research.
Specifically, the retrieved literature $L_{ijk}$ based on embedding similarity is categorized into multiple groups $L_{ijk} = \{n_g\}_{g=1}^G$ according to their publication dates, with each group spanning a period of two years. For each group $g$, the highly cited literature is retained in a top-k manner as the final output for SANA, and the number of literature to be retained for each group is:
\begin{equation}
k_g = \frac{\lvert n_g \lvert}{\lvert L_{ijk} \lvert}K_{O_{ij}},
\end{equation}
where $K_{O_{ij}}$ is a hyper-parameter that represents the number of literature utilized for generating the content of each subsection.

%Reranking is a crucial step that significantly enhances the quality of retrieved information and the relevance to the query. Most existing methods \citep{glass-etal-2022-re2g,bge_embedding} accomplish this by reranking and filtering retrieval results using advanced scoring mechanisms combined with the top-k strategy, ensuring the prioritization of the most relevant and high-quality information for presentation. In the context of survey generation, the quality of the literature is expected to be as high as possible, with citation count regarded as a key indicator of its quality. However, the citation count may be influenced by the publish date of the literature. For instance, earlier-published literature often accumulates more citations over time, while recently published works, despite their potentially higher quality, tend to receive fewer citations. Nevertheless, high-quality recent publications are essential for the content of survey papers. Therefore, the reranking strategy based on time-window is proposed to improve the quality of the retrieved literature. 

\subsubsection{Parallel Generation and Refinement}
Due to the constraints of maximum context length and inference speed of LLMs, the content of each section is generated in parallel to reduce the generation time and ensure the length of the generated survey. However, due to the independent generation processes of each section in parallel, there may be repetition or redundancy among the contents of different section. Therefore, we employ LLMs to implement the refinement stage, which is aimed at refining the raw survey obtained by concatenating the contents of each section generated in parallel.

\subsection{Multi-dimensional Evaluation Benchmark}
\label{sec:benchmark}

% The evaluation of AI-generated surveys remains an open challenge due to the lack of standardized and rigorous benchmarks. Existing methods largely rely on automated scoring by LLMs, face notable limitations: (1) insufficient assessment of whether key literature is adequately covered; (2) reliance on evaluating the full generated content, which may make it difficult for LLMs to capture key aspects, leading to potentially biased or superficial evaluations; (3) dependence solely on the model's internal judgments, lacking objective metrics and expert knowledge to ensure balanced and reliable evaluations.

Evaluating AI-generated surveys is challenging due to the lack of standardized benchmarks. Existing methods largely rely on automated scoring by LLMs, which face limitations: they may not adequately assess key literature coverage and depend heavily on internal model judgments without objective metrics. To address these challenges, we introduce \textbf{SurveyBench}, a comprehensive evaluation benchmark, along with SAM (Survey Assessment Metrics), a multi-dimensional evaluation series. SurveyBench consists of approximately 100 human-written survey papers across 10 distinct topics, carefully curated by doctoral-level researchers to ensure thematic consistency and academic rigor. For each topic $t_i$, we selected one highest-quality survey $S_i^*$ as the reference for comparison with AI-generated surveys $\hat{S}_i$. Details of the benchmark construction process are provided in Appendix.~\ref{details_eval_benchmark}. The SAM series integrate objective metrics, expert knowledge, and multi-dimensional criteria through three core components:

\noindent \textbf{SAM-R: Reference Quality Evaluation.} A comprehensive and relevant bibliography is essential for a well-researched survey. Based on SurveyBench, we extract a reference set $\mathcal{R}_i$ for each topic $t_i$, serving as a reliable benchmark representing foundational knowledge in the field.
% and state-of-the-art 

To measure reference quality, we define the $SAM_R$ metric, which quantifies the overlap between the references in the AI-generated survey $\hat{S}_i$ and $\mathcal{R}_i$:
\begin{equation}
{SAM_R}(\hat{S}_i) = \frac{|R_{{\hat{S}_i}} \cap \mathcal{R}_i|}{|R_{{\hat{S}_i}}|},
\end{equation}
where $R_{{\hat{S}_i}}$ is the set of references in $\hat{S}_i$. A higher rate indicates better coverage of key literature in the topic $t_i$.

\noindent \textbf{SAM-O: Outline Quality Evaluation.}~This component evaluates the structural quality of AI-generated surveys. A well-structured and logically coherent outline is crucial for content organization and readability. We assess the outline using a single comprehensive score $SAM_O$, ranging from 0 to 100, where higher scores indicate better quality. The evaluation is conducted by LLMs following detailed criteria described in Appendix.~\ref{prompt}.

\noindent \textbf{SAM-C: Content Quality Evaluation.} The final component measures the generated survey's quality across three dimensions: structure ($SAM_C^{\text{struct}}$), relevance ($SAM_C^{\text{rel}}$), and coverage ($SAM_C^{\text{cov}}$). Using the high-quality survey $S_i^*$ as reference, we compute avg score of the overall content :

\begin{equation}
SAM_C^{\text{avg}} = \frac{SAM_C^{\text{struct}} + SAM_C^{\text{rel}} + SAM_C^{\text{cov}}}{3}.
\end{equation}

Scores range from 0 to 100, with higher values indicating better performance. The LLMs assess these criteria while referencing $S_i^*$ to ensure alignment with expert-level standards.

%% file: latex/section/exp.tex
\section{Experiment}
\vspace{-0.15cm}
\subsection{Experimental Settings}

\noindent \textbf{Evaluation Dataset.} To assess the performance of our proposed approach, we construct a dedicated benchmark dataset within the Computer Science (CS) domain, based on the \textit{arXiv} repository. As mentioned in Sec.~\ref{sec:benchmark}, we manually select approximately 100 human-written survey papers across 10 distinct topics, and choose one highest-quality survey for direct comparison with AI-generated surveys for each topic.

\begin{table*}
\vspace{-0.5cm}
  \centering
  \resizebox{\textwidth}{!}{
  \begin{tabular}{l|l|cc|c|cccc}
    \hline
    \multirow{2}{*}{\textbf{Methods}} &  \multirow{2}{*}{\textbf{Model}} &\multicolumn{2}{c|}{\textbf{Reference Quality}} & \multirow{2}{*}{\textbf{Outline Quality}} & \multicolumn{4}{c}{\textbf{Content Quality}} \\
    & & Input Cov.  & Reference Cov.  &  & \textbf{Structure} & \textbf{Relevance} & \textbf{Coverage} & \textbf{Avg} \\
    \hline
    Human-Written & - & - & 0.6294 & 87.62 & - & - & - & - \\
    AutoSurvey  & Claude-3-Haiku & 0.1153 & 0.2341 & 82.18 & 72.83 	& 76.44 & 72.35 &73.87  \\
    \textsc{SurveyForge}   & Claude-3-Haiku & 0.2231 & 0.3960 & 86.85 & 73.82 & 79.62 & 75.59 & 76.34 \\
    AutoSurvey  & GPT-4o mini & 0.0665 & 0.2035 & 83.10 & 74.66 & 74.16 & 76.33 & 75.05 \\
    \textsc{SurveyForge}   & GPT-4o mini  & 0.2018 & 0.4236 & 86.62 & 77.10 & 76.94 & 77.15 & 77.06 \\
    \hline
  \end{tabular}
  }
  \vspace{-0.2cm}
  \caption{Comparison of \textsc{SurveyForge} and AutoSurvey~\citep{wang2024autosurvey} using Survey Assessment Metrics (SAM) from three aspects: Reference (SAM-R), Outline (SAM-O) and Content quality (SAM-C). "Input Cov." means the coverage of input papers, measuring the overlap between retrieved papers and benchmark references, while "Reference Cov." means the coverage of reference, evaluating the alignment between cited references of the survey and benchmark references.}
\label{tab:main_results}
\vspace{-0.1cm}
\end{table*}

\noindent \textbf{Implementation Details.} To establish a baseline for comparison, we adopt AutoSurvey  \citep{wang2024autosurvey}, a state-of-the-art system for automated survey generation. Furthermore, we collect a large-scale dataset from the CS scientific field of \textit{arXiv}, consisting of approximately 600,000 research papers and 20,000 review articles. We extract the key metadata to construct a retrieval vector database, including titles, abstracts of all papers and outlines of the review articles. To ensure a fair comparison, we align the timeline of our retrieval database with that of AutoSurvey. During the experimental evaluation, we retrieve 1,500 candidate papers for the outline generation stage and 60 relevant papers for each chapter-writing stage, following the same experimental settings as AutoSurvey.

For survey generation, we employ two LLMs independently: \hytt{Claude-3-haiku-20240307} and \hytt{GPT-4o-mini-2024-07-18}. Each model generates surveys for 10 predefined topics, with 10 independent trials conducted for each topic, resulting in a total of 100 outputs per model. The average performance across these trials is calculated to ensure stable and reliable results. In addition to the closed-source models, we have also experimented with the open source model with Deepseek-v3~\cite{liu2024deepseek}, with impressive results, as detailed in Appendix~\ref{sec:opensource_exp}. For evaluation, we leverage more advanced models, \hytt{GPT-4o-2024-08-06} and \hytt{Claude-3.5-sonnet-20241022}, to assess both the AI-generated outlines and the content of the surveys, ensuring a robust and reliable evaluation of their quality.

\begin{table*}
  \centering
  \resizebox{0.94\textwidth}{!}{
  \begin{tabular}{l|ccc | cc}
    \hline
    \multirow{2}{*}{\textbf{Methods}} & \multicolumn{3}{c|}{\textbf{Outline Comparison}} & \multicolumn{2}{c}{\textbf{Content Comparison}}\\
    & Score Win Rate & Comparative Win Rate & Human Eval & Score Win Rate &  Human Eval \\
    \hline
    AutoSurvey~\citep{wang2024autosurvey}  & 27.00\% & 25.00\% & 26.00\% & 31.00\% & 30.00\% \\
    \textsc{SurveyForge}        & 73.00\%  & 75.00\% & 74.00\% & 69.00\% & 70.00\% \\
    \hline
  \end{tabular}
  }
  \vspace{-0.2cm}
  \caption{Win-rate comparison of automatic and human evaluations on outline and content quality. "Score Win Rate" reflects the win rate based on individual LLM-scores, where the LLM assigns separate score to each survey paper before determining the higher-scoring one. "Comparative Win Rate" is derived from LLM pairwise comparisons, where the LLM directly compares two articles side-by-side and decides which one is superior. "Human Eval" represents the win rate derived from expert human evaluations.}
  \label{tab:win_rate}
\end{table*}

\subsection{Main Results}
As shown in Table~\ref{tab:main_results}, we evaluate the performance of \textsc{SurveyForge} across various dimensions, including reference quality, outline quality, and content quality, comparing it against the baseline AutoSurvey. The results demonstrate that \textsc{SurveyForge} achieves significant improvements in all aspects, showcasing its potential as an advanced automated survey generation framework. Additionally, we conduct a cost analysis of the \textsc{SurveyForge} framework, demonstrating that generating a 64k-token overview requires less than \textdollar 0.50, with detailed cost breakdowns provided in Appendix~\ref{framwork_cost}.

\begin{figure}[t!]
\vspace{-0.25cm}
    \centering
    \begin{subfigure}[b]{0.23\textwidth}
        \includegraphics[width=\textwidth]{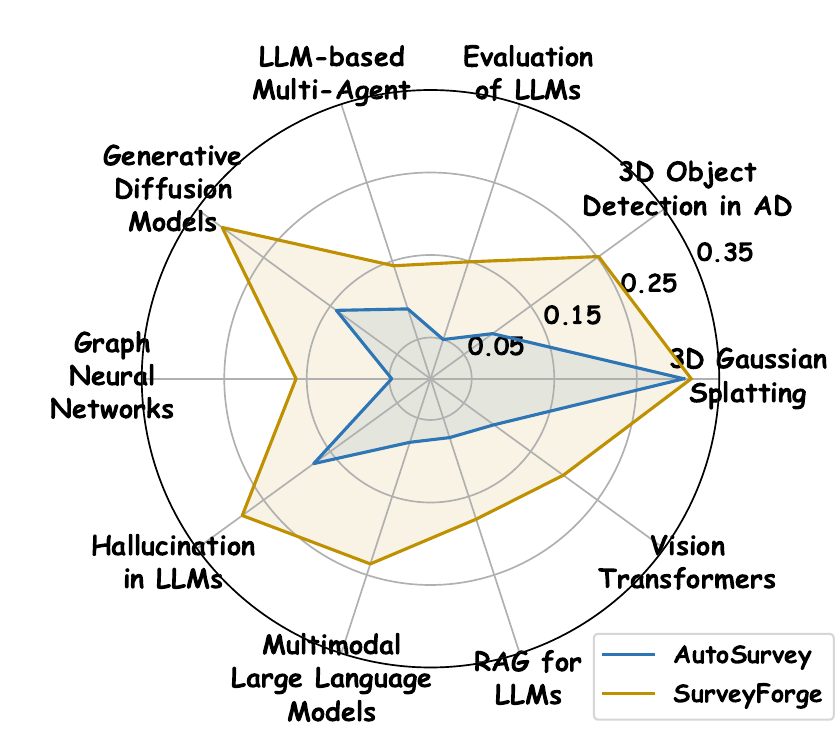}
        \caption{}
        \label{fig:rag}
    \end{subfigure}
    \hfill
    \begin{subfigure}[b]{0.23\textwidth}
        \includegraphics[width=\textwidth]{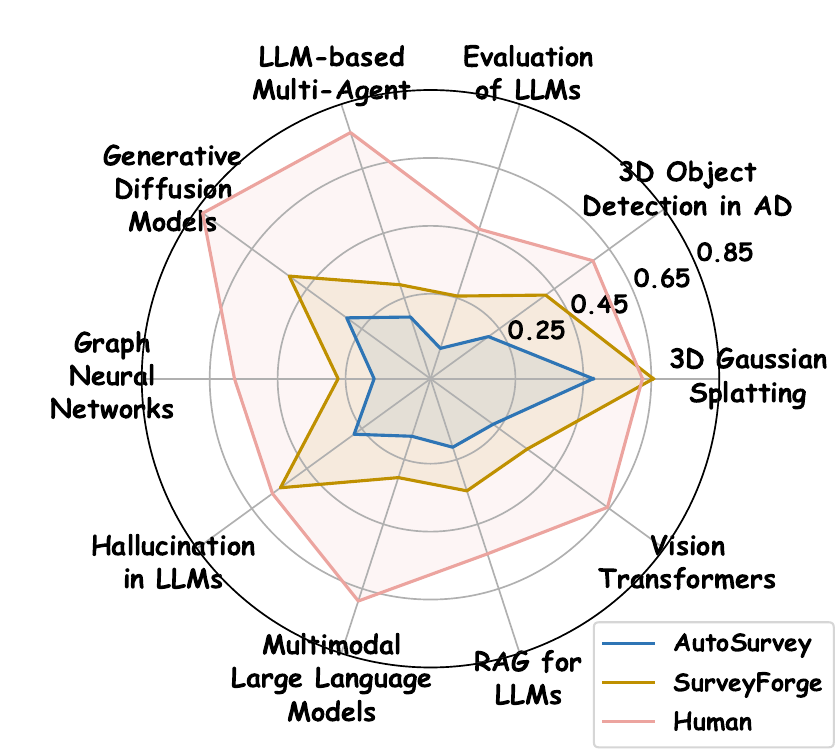}
        \caption{}
        \label{fig:ref}
    \end{subfigure}
    
    \begin{subfigure}[b]{0.23\textwidth}
        \includegraphics[width=\textwidth]{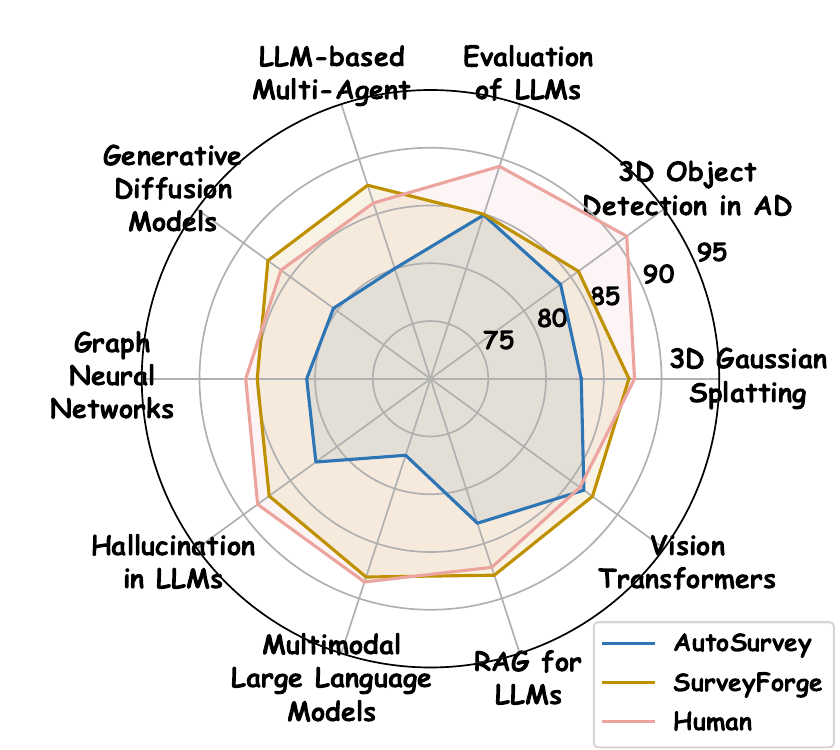}
        \caption{}
        \label{fig:outline}
    \end{subfigure}
    \hfill
    \hspace{0.005\textwidth}
    \begin{subfigure}[b]{0.23\textwidth}
        \includegraphics[width=\textwidth]{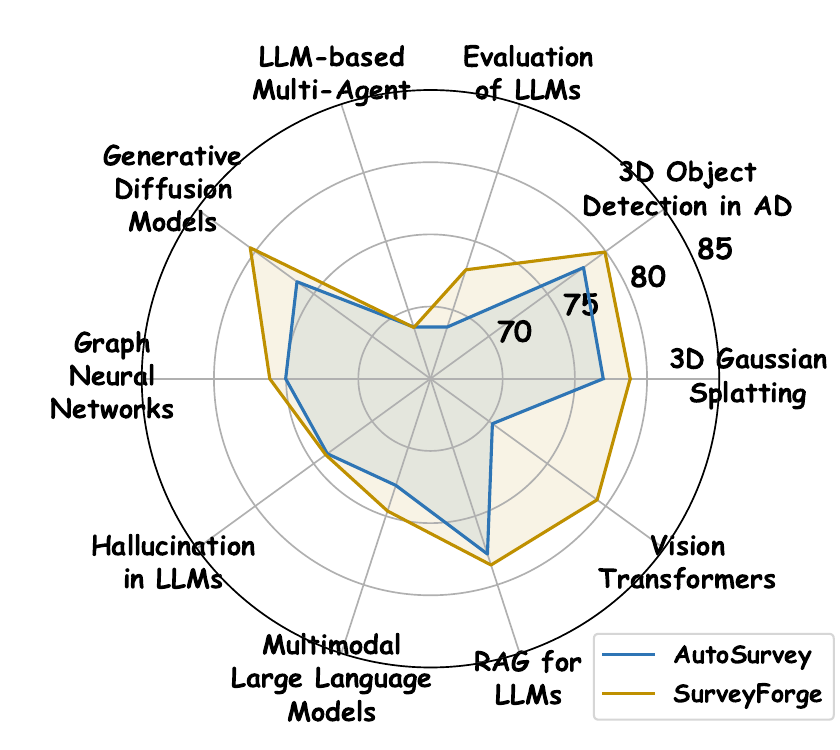}
        \caption{}
        \label{fig:content}
    \end{subfigure}
    \vspace{-0.70cm}
    \caption{Evaluation results on SurveyBench. Evaluation results of (a) Input Coverage, (b) Reference Coverage, (c) Outline Quality, and (d) Content Quality.}
    \label{fig:all_results}
\vspace{-0.5cm}
\end{figure}

\noindent \textbf{Results on Reference Quality.} In terms of reference quality, \textsc{SurveyForge} outperforms AutoSurvey on both key metrics: Input Coverage, which measures the relevance of retrieved papers, and Reference Coverage, which evaluates the alignment of the references of surveys with expert-curated benchmarks. Specifically, the Input Coverage score improves from 0.12 to 0.22 when using Claude-3-Haiku and from 0.07 to 0.20 with GPT-4o mini. Similarly, the Reference Coverage score increases from 0.23 to 0.40 and from 0.20 to 0.42 for the two respective models, indicating that \textsc{SurveyForge} retrieves and generates references that are not only more relevant but also more aligned with expert expectations. Notably, high-quality human-written surveys achieve a Reference Coverage score of 0.63, which further validates the reliability of our proposed reference evaluation database, which provides a robust benchmark for reference quality.

\noindent \textbf{Results on Outline Quality.} For outline quality, the results show that \textsc{SurveyForge} generates outlines that are more logical, comprehensive, and closer to human-level performance compared to AutoSurvey \citep{wang2024autosurvey}. Using Claude-3-Haiku, the outline quality score increases from 82.25 to 86.58, while GPT-4o mini achieves a similar improvement from 83.10 to 86.62. These advancements are driven by the proposed few-shot heuristic learning method, which leverages expert-curated examples from the Survey Outline Database to guide the LLMs in producing well-structured and domain-relevant outlines.

\begin{table}[t]
\vspace{-0.15cm}
  \centering
  \resizebox{0.49\textwidth}{!}{
  \begin{tabular}{l|c|c|c}
    \hline
    \textbf{Method} & \textbf{Heuristic Learning} & \textbf{Demonstration Outline} & \textbf{Outline Quality} \\
    \hline
    AutoSurvey & $\times$ & - & 81.78 \\
    \textsc{SurveyForge} & \checkmark & From random surveys & 84.58 \\
    \textsc{SurveyForge} & \checkmark & From related surveys & 86.67 \\
    \hline
  \end{tabular}
  }
  \vspace{-0.2cm}
  \caption{Ablation study for outline generation. "Demonstration Outline" means the source of outlines used for heuristic learning.}
  \label{tab:abl_outline}
\vspace{-0.4cm}
\end{table}

% \begin{table}[t]
%   \centering
%   \resizebox{0.36\textwidth}{!}{
%   \begin{tabular}{l|c}
%     \hline
%     \textbf{Methods} & \textbf{Outline Quality} \\
%     \hline
%     AutoSurvey        & 81.78 \\
%     SurveyForge (RH)  & 84.58 \\
%     SurveyForge      & 86.67 \\
%     \hline
%   \end{tabular}
%   }
%   \vspace{-0.2cm}
%   \caption{Ablation study for outline generation. SurveyForge (RH) represents random, non-domain-specific outlines are used as structural guidance during heuristic learning, SurveyForge means using domain-related outlines retrieved.}
%   \label{tab:abl_outline}
% \vspace{-0.4cm}
% \end{table}

\noindent \textbf{Results on Content Quality.} For content quality, \textsc{SurveyForge} achieves consistent improvements across all three evaluation dimensions: structure, relevance, and coverage. The average content quality score increases from 73.87 to 76.34 (Claude-3-Haiku) and 75.05 to 77.06 (GPT-4o mini). These results confirm that \textsc{SurveyForge} generates content that is better organized, more relevant, and more comprehensive, effectively addressing the critical aspects of the target domain.

As shown in Fig.~\ref{fig:all_results}, \textsc{SurveyForge} demonstrates substantial improvements over the baseline AutoSurvey across all key evaluation metrics. Although not yet matching the quality of expert-crafted surveys, \textsc{SurveyForge} significantly narrows the gap, highlighting its potential as a powerful tool for automated survey generation.
% Although there is still some gap compared to high-quality surveys written by human experts

\subsection{Comparison with Human Evaluation}
To validate our automated evaluation system, we compare its performance with expert assessments using 100 outputs from \hytt{Claude-3-haiku-20240307} across 10 topics (Please refer to Appendix~\ref{details_eval_benchmark} and Appendix~\ref{details_human_eval} for detail information). We employ a win rate framework, presenting the anonymized results of \textsc{SurveyForge} and AutoSurvey~\cite{wang2024autosurvey} to 20 PhD experts in computer science field. These experts were carefully selected according to the evaluation topic and processes deep expertise in the relevant domain. 
%Concurrently, the automated system calculated two metrics: Score Win Rate, based on LLM-assigned scores, and Comparative Win Rate, from direct pairwise comparisons by LLMs.

As shown in Table~\ref{tab:win_rate}, for outline quality, the automated system achieves a Score Win Rate of 73.00\% and a Comparative Win Rate of 75.00\%, closely matching the human evaluation rate of 74.00\%. This consistency confirms the system's robust scoring logic. For content quality, the automated system's Score Win Rate for \textsc{SurveyForge} is 69.00\%, aligning closely with the human expert rate of 70.00\%. In addition, we also conduct Cohen's kappa coefficient consistency experiment, which shows a strong agreement between automated systems and human assessments, as detailed in Appendix~\ref{details_human_eval}.

In summary, the automated system aligns well with human assessments for both outline and content quality, validating its effectiveness as a reliable alternative to manual evaluation.

\begin{table}[t]
  \resizebox{0.94\linewidth}{!}{
  \centering
  \begin{tabular}{ccc|cc}
    \hline
    \multicolumn{3}{c|}{\textbf{Components}} & \multicolumn{2}{c}{\textbf{Reference Quality}} \\
    MR & MS & TRE & Input Cov. & Reference Cov. \\
    \hline
    -& -& -& 0.1119 & 0.2340 \\
    \checkmark & -& -& 0.1694& 0.2730\\
    \checkmark & \checkmark & -&  0.1781&  0.2984\\
    \checkmark & -& \checkmark &  0.1997&  0.3542\\
    \checkmark & \checkmark & \checkmark &   0.2224 & 0.3971 \\
    \hline
  \end{tabular}
  }
  \vspace{-0.2cm}
  \caption{Ablation study for content generation. We perform ablation on three components of SANA module: MR represents Memory for Retrieval, MS represents Memory for Sub-query, and TRE represents Temporal-aware Reranking Engine.}
  \label{tab:abl_rag}
\vspace{-0.3cm}
\end{table}

\subsection{Ablation Study}
To better understand the contribution of individual components in our proposed \textsc{SurveyForge} framework, we conduct a comprehensive ablation study. For ablation experiments, we use \hytt{Claude-3-haiku-20240307} to generate surveys on the same 10 topics, with 3 independent trials per topic to ensure statistical reliability while maintaining computational efficiency. Specifically, we analyze the memory mechanism, sub-query decomposition, and reranking strategies in the scholar navigation agent module, as well as the impact of the use of the database of survey outlines in the outline generation process. The results of the ablation experiments are presented in Table~\ref{tab:abl_outline} and Table~\ref{tab:abl_rag}.

\noindent \textbf{Analysis on Outline Generation.} Table~\ref{tab:abl_outline} highlights the impact of heuristic learning approach on outline quality. The baseline method, which generates outlines solely from retrieved research papers without structural guidance, achieves a score of 81.78. This indicates the absence of organizational cues limits the coherence and logical flow of the outlines. To address this, we first introduce a heuristic approach using outlines from random surveys. These generic outlines, representing common patterns in survey writing, improve the score to 84.58. This shows the effectiveness of structural cues, even without target-domain tailoring. Finally, we retrieve domain-specific outlines, providing both structural guidance and thematic alignment with the target domain. As a result, the outline quality score significantly rises to 86.67, showing the crucial role of domain-specific structural cues in creating coherent and relevant outlines.

% even when they are not tailored to the target domain

\noindent \textbf{Analysis on Content Generation.} 
Based on the experimental results presented in Table ~\ref{tab:abl_rag}, it can be observed that as the quality of literature obtained by SANA improves, the quality of cited references in surveys also correspondingly enhances. This observation highlights the importance of using SANA during the content generation stage to retrieve high-quality literature. Specifically, the integration of a memory mechanism into the query decomposition and retrieval processes significantly enhance the quality of literature. This improvement can be attributed to the incorporation of more comprehensive sub-query semantics and a retrieval scope better aligned with the sub-queries. Besides, the temporal-aware reranking engine ensures the selection of high-quality papers, leading to a more comprehensive and balanced reference collection.

%% file: latex/section/conclusion.tex
\section{Conclusion and Outlook} 
\vspace{-0.15cm}
We have introduced \textsc{SurveyForge}, an automated framework leveraging a heuristic outline generation and a memory-driven content generation to generate high-quality surveys. We introduce a multi-dimensional evaluation benchmark to comprehensively assess the quality of surveys. \textsc{SurveyForge} significantly outperforms prior approaches across multiple evaluation metrics. We hope to reduce the learning curve for researchers venturing into unfamiliar fields, providing convenience and thereby promoting the integration and development of cross-disciplinary and cross-domain knowledge.

% In future research, we hope to enhance \textsc{SurveyForge} to facilitate the progress of the automatic scientific research project. For instance, by leveraging \textsc{SurveyForge}, we can quickly generate academic surveys of research papers on a specific issue or task, and automatically analyze and extract academic perspectives that can be adopted from the surveys. This will reduce the learning curve for researchers venturing into unfamiliar fields, providing convenience and thereby promoting the integration and development of cross-disciplinary and cross-domain knowledge.

% addressing key challenges in existing methods. 

% covering outline, content, and citation quality. Experimental results demonstrate that 

% Looking ahead, we aim to enhance  to facilitate the integration of cross-disciplinary knowledge, thereby fostering the convergence of diverse disciplinary ideas to drive the automation of scientific research.

%% file: latex/section/Limitations.tex
\section*{Limitations}
Despite its strong performance in generating structured and high-quality surveys, \textsc{SurveyForge} has inherent limitations, as discussed in Appendix~\ref{sec:discussion_ai_human}. While LLMs excel at summarizing existing literature, they face challenges in analyzing and synthesizing relationships across multiple sources, often lacking the critical thinking and originality characteristic of human-authored work, which limits their capability to reflect research trends or provide forward-looking insights. Besides, the accuracy of content and citations is also affected by the hallucination of LLMs. Future work could focus on developing methods to better capture interconnections among references to enhance the logical coherence, depth, and scholarly value of the generated content.

%% file: latex/section/supp.tex
\clearpage
\newpage
\setcounter{page}{1}
\appendix

\section{Appendix}
Due to the page limitation of the manuscript, we provide more details and visualizations from the following aspects:

\begin{itemize}
    \item Sec.~\ref{details_databse}: Database Construction.
    \item Sec.~\ref{details_eval_benchmark}: Details of SurveyBench.
    \item Sec.~\ref{sec:discussion_ai_human}: Discussion about Generated Surveys and Human-written Surveys.
    \item Sec.~\ref{details_human_eval}: Details of Human Evaluation and Inter-rater Agreement.
    \item Sec.~\ref{sec:opensource_exp}: Additional Experiments with Open-Source Models.
    \item Sec.~\ref{framwork_cost}: Details of Time and Economic Cost.
    \item Sec.~\ref{vis_outline}: Qualitative Results.
    \item Sec.~\ref{vis_survey}: Example of Generated Survey.
    \item Sec.~\ref{prompt}: Prompt Used.
\end{itemize}

\subsection{Database Construction}
\label{details_databse}

To ensure the quality and relevance of the AI-generated surveys, we construct two key databases: the \textit{Research Paper Database} and the \textit{Survey Outline Database}, consisting of approximately 600,000 research papers and 20,000 review articles, which together serve as the foundation for content generation and structural guidance. The \textit{Research Paper Database} comprises the titles and abstracts of research papers relevant to the survey topic, while the \textit{Survey Outline Database} contains titles, abstracts, and outlines extracted from published survey papers. 

Specifically, we utilize MinerU \citep{wang2024mineru} to extract content from a corpus of survey articles. Using rule-based extraction techniques, we isolate hierarchical outlines, including section and subsection headings. However, due to variations in formatting and structure across different papers, automatic extraction may introduce noise. To address this, we employ \hytt{Claude-3.5-sonnet-20241022} to refine and standardize the extracted outlines, ensuring consistency in structure and formatting. By leveraging the \textit{Survey Outline Database} in this way, we provide the LLM with high-quality, expert-crafted outline examples to guide its generation process. 

Additionally, we encode these documents using the \hytt{gte-large-en-v1.5} embedding model \citep{li2023towards}, which captures semantic relationships and enables efficient similarity-based retrieval. This combination of structured expert examples and semantic encoding ensures a robust foundation for outline generation and content retrieval.

\begin{table*}[h]
\vspace{-0.25cm}
\centering
\renewcommand{\arraystretch}{1.45}
\resizebox{\textwidth}{!}{
\begin{tabular}{l|clc}
\hline
\textbf{Topic} & \textbf{Ref Num} & \textbf{Selected Survey Title} & \textbf{Citation} \\ \hline
Multimodal Large Language Models & 912 & A Survey on Multimodal Large Language Models & 979 \\ \hline
Evaluation of Large Language Models & 714 & A Survey on Evaluation of Large Language Models & 1690 \\ \hline
3D Object Detection in Autonomous Driving & 441 & 3D Object Detection for Autonomous Driving: A Comprehensive Survey & 172 \\ \hline
Vision Transformers & 563 & A Survey of Visual Transformers & 405 \\ \hline
Hallucination in Large Language Models & 500 & Siren's Song in the AI Ocean: A Survey on Hallucination in Large Language Models & 808 \\ \hline
Generative Diffusion Models & 994 & A Survey on Generative Diffusion Models & 367 \\ \hline
3D Gaussian Splatting & 330 & A Survey on 3D Gaussian Splatting & 128 \\ \hline
LLM-based Multi-Agent & 823 & A Survey on Large Language Model Based Autonomous Agents & 765 \\ \hline
Graph Neural Networks & 670 & Graph Neural Networks: Taxonomy, Advances, and Trends & 129 \\ \hline
Retrieval-Augmented Generation for Large Language Models & 608 & Retrieval-Augmented Generation for Large Language Models: A Survey & 953 \\ \hline
\end{tabular}
}
\vspace{-0.25cm}
\caption{Overview of selected topics and the representative surveys in our evaluation benchmark. For each topic, we show the total number of unique references (Ref Num) collected from SurveyBench, and the citation count of selected high-quality surveys that serve as our evaluation references.}
\label{tab:survey_topics}
\vspace{-0.25cm}
\end{table*}

\subsection{Details of SurveyBench}
\label{details_eval_benchmark}

To construct SurveyBench, we select 10 trending topics in the computer science domain, as shown in Table~\ref{tab:survey_topics}. These topics span various cutting-edge areas including multimodal learning, language models, computer vision, and autonomous systems. For each topic, a set of high-quality, human-written surveys is carefully curated by \textit{a panel of 20 researchers}. Each of these researchers holds doctoral degrees and possesses extensive expertise in the aforementioned 10 trending topics in the computer science domain. This rigorous selection process ensures strong thematic alignment and guarantees the inclusion of authoritative and relevant surveys. Besides, the development of our assessment metrics (e.g. SAM-O and SAM-C) is inspired by peer review guidelines from top-tier computer science venues. However, we observed that traditional review criteria often rely heavily on reviewers' implicit knowledge and experience, making them challenging to implement in automated evaluation systems. To address this limitation, we systematically decomposed these high-level review guidelines into more specific, measurable components that can be reliably assessed by LLMs while maintaining consistency with expert human evaluation. For example, in our outline assessment criteria, abstract concepts like "topic organization" were broken down into concrete, assessable elements such as "topic uniqueness" (checking for duplicate topics, content overlap) and "structural balance" (examining section development and proportionality). This granular approach, developed through discussions with researchers who have at least two years of reviewing experience for top CS venues, enables more consistent and reliable automated evaluation across different survey topics while preserving the essential quality standards of academic peer review.

The curated surveys, predominantly published within the last two years, are chosen to ensure both timeliness and relevance. From each selected survey, we extract the references cited to construct a dedicated reference database for each topic, resulting in comprehensive reference collections ranging from 330 to 994 references per topic, as detailed in Table~\ref{tab:survey_topics}. Furthermore, to facilitate robust content evaluation,we identify the highest-quality survey for each topic to serve as the evaluation reference, with these selected surveys demonstrating significant impact through their citation counts (ranging from 128 to 1,690 citations). SurveyBench provides a comprehensive and reliable foundation for assessing the quality of AI-generated surveys, ensuring both reference coverage and content relevance are rigorously evaluated.

\subsection{Discussion about Generated Surveys and Human-written Surveys}
\label{sec:discussion_ai_human}

While our extensive evaluation of \textsc{SurveyForge} demonstrates its effectiveness in automated survey generation, our analysis reveals several fundamental challenges that warrant further investigation. Through systematic examination of the generated surveys, we identify two primary limitations of the current system.

The first limitation lies in the depth of academic analysis. Although the system effectively extracts and organizes information from individual papers, it exhibits constraints in establishing profound connections across multiple publications. Specifically, the system's capability falls short in comparative analysis of temporal innovations and methodological evolution patterns, often defaulting to mechanical reference listing rather than providing the nuanced synthesis characteristic of expert-written surveys. This limitation stems primarily from challenges in the accurate identification of the core literature and the construction of deep logical relationships during the processing of long-form knowledge.

The second challenge concerns the accuracy of content and citation. Despite our implementation of multiple verification mechanisms, the system occasionally produces inaccurate citations or academic claims, potentially affecting the survey's reliability. This remains a critical area for improvement in automated survey generation systems.

To address these limitations, future work could focus on developing comprehensive knowledge association networks through core entity extraction and citation graph construction, which may enhance the system's capability to identify deep inter-publication connections.

\subsection{Details of Human Evaluation and Inter-rater Agreement}
\label{details_human_eval}

\begin{table}[t]
\vspace{-0.15cm}
  \centering
  \setlength\tabcolsep{18pt}
  \resizebox{0.49\textwidth}{!}{
  \begin{tabular}{l|c|c}
    \hline
    \textbf{Evaluation Pair} & \textbf{Aspect} & \textbf{$\kappa$} \\
    \hline
    LLM vs. Human & Outline & 0.7177 \\
    LLM vs. Human & Content & 0.6462 \\
    Human Cross-Validation & Outline & 0.7921 \\
    Human Cross-Validation & Content & 0.7098 \\
    \hline
  \end{tabular}
  }
  \vspace{-0.2cm}
  \caption{Inter-rater agreement between LLM and human evaluations. \textbf{$\kappa$} means the Cohen's kappa coefficient.}
  \label{tab:inter-rater}
\vspace{-0.4cm}
\end{table}

For the human evaluation across the selected 10 topics, we recruited 20 PhD experts in computer science from various prestigious institutions, including several QS Top 50 universities and renowned research institutes within our country. The selection of these experts followed strict criteria to ensure their expertise and qualifications. All evaluators hold PhD degrees in computer science or closely related fields, and each expert has published at least one peer-reviewed paper in the specific topic they were assigned to evaluate. Moreover, all selected experts are currently active researchers in their respective fields.

To maintain evaluation quality and consistency, each expert was provided with a comprehensive evaluation guideline manual, identical to the one used in our LLM evaluation system, ensuring consistent assessment criteria across all evaluators. Before the formal evaluation, we conducted a training session to familiarize the experts with the evaluation criteria and scoring rubrics. The evaluation process was conducted in a double-blind manner to minimize potential biases. Regarding compensation, experts were paid \$50 per hour, commensurate with their expertise level. The average evaluation time per survey was approximately 1-3 hours, ensuring thorough and reliable assessment.

To further verify the reliability of the evaluation system, we further conducted Cohen's kappa coefficient experiment to measure the inter-rater agreement between automatic and human evaluations and evaluations inter-rater agreement among human annotators. Specifically, as shown in Table~\ref{tab:inter-rater}, we conducted a systematic evaluation of 100 generated survey papers across 10 different research topics. We used Cohen's kappa coefficient as our evaluation metric, covering two core dimensions: outline and content.

In the outline dimension, based on the evaluation of these 100 surveys, the kappa coefficient between LLM evaluation and human evaluation reached 0.7177, indicating significant agreement between the two. Meanwhile, the cross-validation kappa coefficient between human evaluators was 0.7921. This high level of agreement not only validates the reliability of human evaluation but also supports the effectiveness of our automated evaluation method.

In the content dimension, based on the same sample size, the kappa coefficient between LLM evaluation and human evaluation was 0.6462, while the cross-validation kappa coefficient between human evaluators was 0.7098. These results demonstrate that even in the more complex task of evaluating extra-long text content, our evaluation framework still shows good consistency.

\subsection{Additional Experiments with Open-Source Models}
\label{sec:opensource_exp}

To validate the generalizability of our framework, we conduct additional experiments using DeepSeek-v3~\cite{liu2024deepseek}, a state-of-the-art open-source language model. As shown in Table~\ref{tab:comp_open_closed}, the experimental results demonstrate remarkable performance across all evaluation metrics. Specifically, DeepSeek-v3 achieved an Input Coverage of 0.2554 and a Reference Coverage of 0.4553, surpassing other baseline models in literature coverage assessment. In the outline quality evaluation, DeepSeek-v3 attains a score of 87.42, which not only exceeds other models but also approaches the benchmark set by human-written surveys (87.62). Furthermore, across the three dimensions of content quality structure, relevance, and coverage, DeepSeek-v3 demonstrates exceptional performance with scores of 79.20, 80.17, and 81.07 respectively, yielding a mean score of 80.15 that outperforms other comparative models.

These empirical results not only corroborate the effectiveness of our methodology but also establish its applicability to open-source models. Notably, DeepSeek-v3~\cite{liu2024deepseek} exhibits superior performance at a lower operational cost (\textdollar 0.37 per survey) compared to GPT-4o-mini (\textdollar 0.43 per survey). Such advancement has substantial implications for the sustainable development of automated research tools and methodologies.

\begin{table*}
\vspace{-0.5cm}
  \centering
  \resizebox{\textwidth}{!}{
  \begin{tabular}{l|l|cc|c|cccc}
    \hline
    \multirow{2}{*}{\textbf{Methods}} &  \multirow{2}{*}{\textbf{Model}} &\multicolumn{2}{c|}{\textbf{Reference Quality}} & \multirow{2}{*}{\textbf{Outline Quality}} & \multicolumn{4}{c}{\textbf{Content Quality}} \\
    & & Input Cov.  & Reference Cov.  &  & \textbf{Structure} & \textbf{Relevance} & \textbf{Coverage} & \textbf{Avg} \\
    \hline
    Human-Written & - & - & 0.6294 & 87.62 & - & - & - & - \\
    \textsc{SurveyForge}   & Claude-3-Haiku & 0.2231 & 0.3960 & 86.85 & 73.82 & 79.62 & 75.59 & 76.34 \\
    \textsc{SurveyForge}   & GPT-4o mini  & 0.2018 & 0.4236 & 86.62 & 77.10 & 76.94 & 77.15 & 77.06 \\
    \textsc{SurveyForge} & Deepseek-v3 & 0.2554	& 0.4553 & 87.42 &	79.20	& 80.17	& 81.07	& 80.15 \\
    \hline
  \end{tabular}
  }
  \vspace{-0.2cm}
  \caption{Comparison of open source and closed source models on SurveyBench.}
\label{tab:comp_open_closed}
\vspace{-0.1cm}
\end{table*}

\subsection{Details of Time and Economic Cost}
\label{framwork_cost}
The \textsc{SurveyForge} framework generates comprehensive survey papers with approximately 64k tokens in length, comparable to human-written surveys. The generation process requires an average input of 2.37M tokens and produces 0.13M tokens of output. Taking GPT-4-mini-2024-07-18 as an example, the economic cost amounts to merely \textdollar 0.43. Regarding the temporal efficiency, the entire framework completes the generation within approximately 10 minutes (note that the actual duration may vary depending on API rate limits). These metrics demonstrate that the \textsc{SurveyForge} framework enables researchers to efficiently acquire domain knowledge at a remarkably low cost.

\subsection{Qualitative Results}
\label{vis_outline}

\begin{figure*}[t!]
   \vspace{-0.4cm}
   \includegraphics[width=\linewidth]{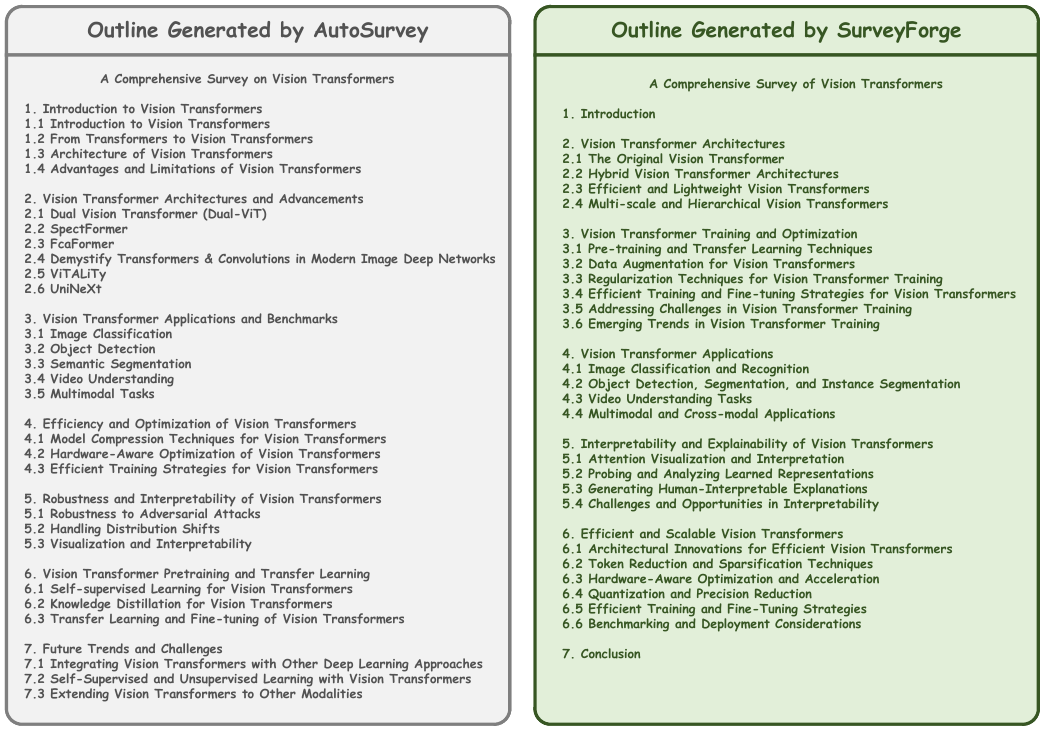}
   \vspace{-0.2cm}
   \includegraphics[width=\linewidth]{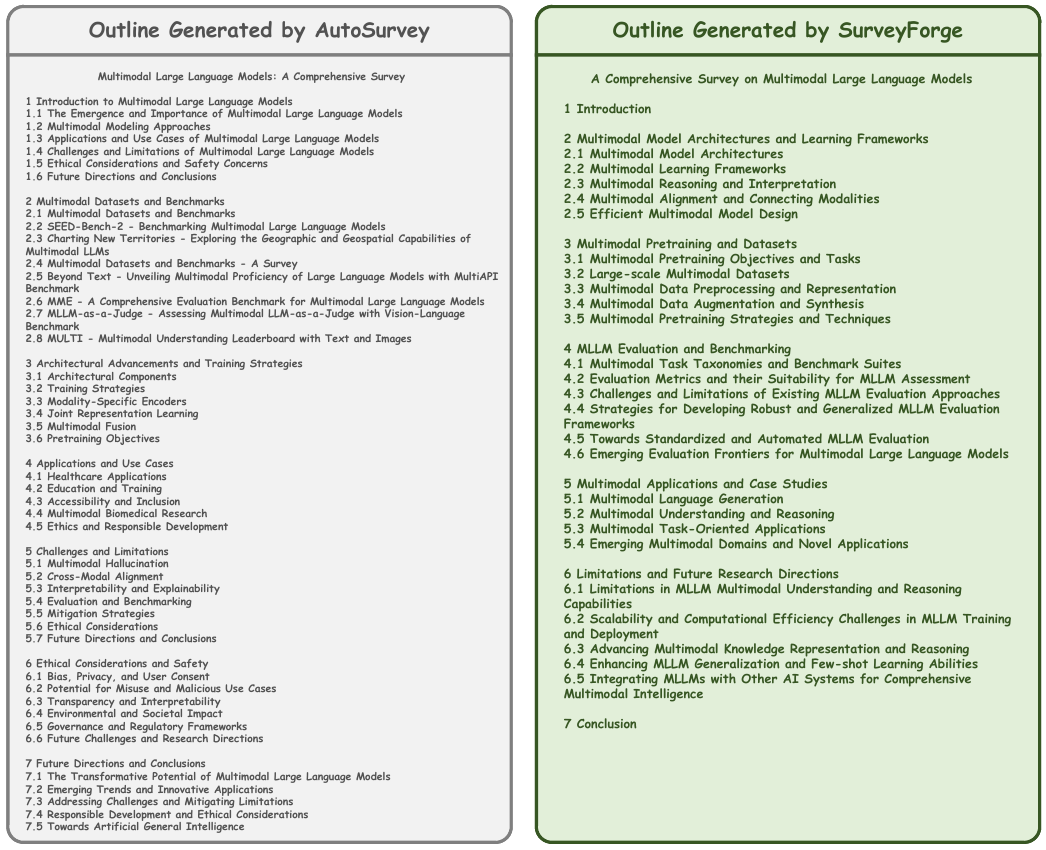}
   % \vspace{-0.3cm}
   \caption{Comparisons of survey outlines generated by the baseline method (left) and our proposed framework (right). The baseline displays a fragmented structure, whereas our method yields a more comprehensive, systematically organized outline.}
   \label{fig:outline_comparison}
\end{figure*}

In this section, we present qualitative comparisons to demonstrate the effectiveness of our proposed framework in generating academically structured survey outlines. Specifically, we compare the outlines generated by our method with those produced by baseline approaches, as shown in Fig.~\ref{fig:outline_comparison}.

The baseline outlines exhibit several notable issues. First, the logical organization of sections and subsections is often suboptimal, with limited hierarchical depth and coherence. Additionally, there is a tendency to treat individual studies or papers as standalone subsections, resulting in fragmented and overly granular structures. Furthermore, redundancy is frequently observed, with similar or overlapping topics appearing in multiple sections, which reduces clarity and disrupts the logical flow of the outline.

In contrast, the outlines generated by our framework effectively address these issues. By leveraging a heuristic learning approach and incorporating domain-specific structural patterns, our method produces well-organized outlines that align with academic writing standards. The generated outlines demonstrate clear hierarchical organization, thematic coherence, and appropriate grouping of related topics, providing a solid foundation for comprehensive and logically structured surveys.

\subsection{Example of Generated Survey}
\label{vis_survey}
\begin{figure*}[t!]
   \centering
   \includegraphics[width=0.92\linewidth]{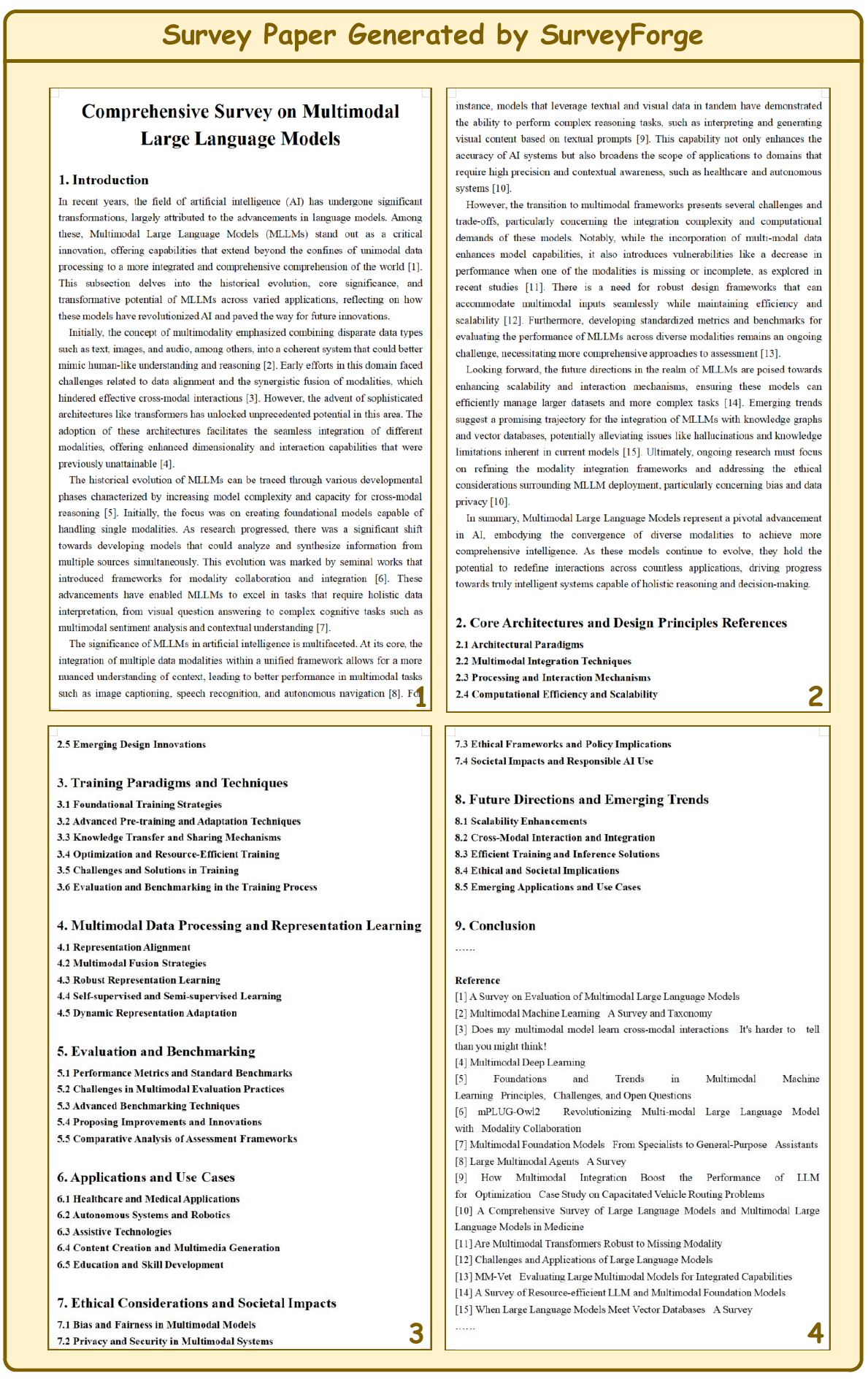}
   \caption{Example of the survey generated by \textsc{SurveyForge}. Please refer to~\textcolor{teal}{\url{https://anonymous.4open.science/r/survey_example-7C37/}} for more auto-generated results.}
   \label{fig:survey_vis}
\end{figure*}

As shown in Fig.~\ref{fig:survey_vis}, we have provided the example of the generated survey by \textsc{SurveyForge}, more complete examples can be found at~\textcolor{teal}{\url{https://anonymous.4open.science/r/survey_example-7C37/}}. Specifically, by observing the generated survey paper, we found that \textsc{SurveyForge} is not only capable of summarizing knowledge within a specific academic field based on logical structures but also excels at providing insights and recommendations for some potential research directions. 

For instance, in a survey paper generated by \textsc{SurveyForge} titled "Comprehensive Survey on Multimodal Large Language Models: Advances, Challenges, and Future Directions", Section 8 offers a detailed outlook on several potential future technological pathways for Multimodal Large Language Models (MLLMs), such as scalability enhancements, cross-modal interaction and integration, and efficient training and inference solutions. Besides, the survey paper also raises concerns about the ethical and societal implications of the excessive use of MLLMs, including their potential impact on issues such as gender, race, ethnicity, and socioeconomic status. Furthermore, \textsc{SurveyForge} has outlined numerous application scenarios for MLLMs, including AI-driven agents, interactive systems, Augmented Reality (AR), and specialized domains such as healthcare and education. In addition, \textsc{SurveyForge} further analyzes the challenges that need to be addressed to apply MLLMs to these practical scenarios. For instance, addressing computational limitations and tackling privacy concerns associated with systems that rely on large amounts of data, which require robust frameworks for data management and obtaining user consent.

\subsection{Prompt Used}
\label{prompt}
This section outlines the key prompts employed in \textsc{SurveyForge}, covering those for outline generation, content generation, and evaluation.

The outline generation prompt incorporates two key elements: the structure of human-written survey papers and relevant literature on the topic. This prompt ensures that the generated outline adheres to academic conventions, with section titles aligned to the survey topic, maintaining logical connections between sections while avoiding redundancy. The content generation prompt guides LLMs in drafting individual sections of a survey paper. It requires the generated content to be supported by references from relevant literature and specifies length constraints to ensure clarity and precision.

For the prompts used for evaluation, we design the evaluation rules from both the outline and the content. Regarding outline evaluation, LLMs are instructed to score from the aspects of topic uniqueness, structural balance, hierarchical clarity and logical organization, with the total score for each aspect serving as the overall score for the outline. For content evaluation, the process references human-written surveys: LLMs first review such surveys on the same topic to establish context before evaluating AI-generated content. This approach grounds the evaluation in established academic writing practices, enhancing the reliability of the assessment.

\begin{figure*}[t!]
     \centering
     \includegraphics[width=0.99\linewidth]{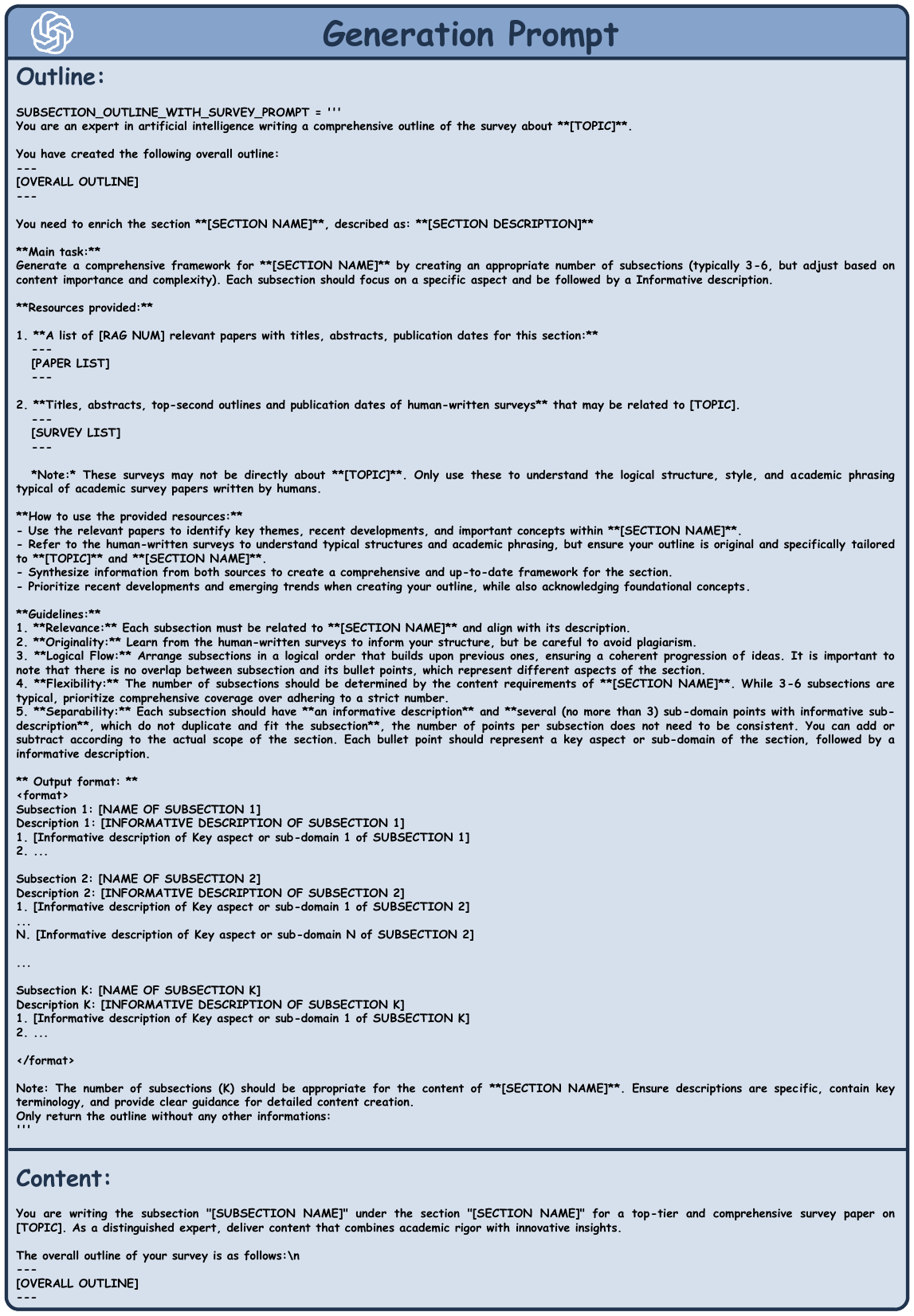}
\end{figure*}

\begin{figure*}[t!]
     \centering
     \includegraphics[width=0.99\linewidth]{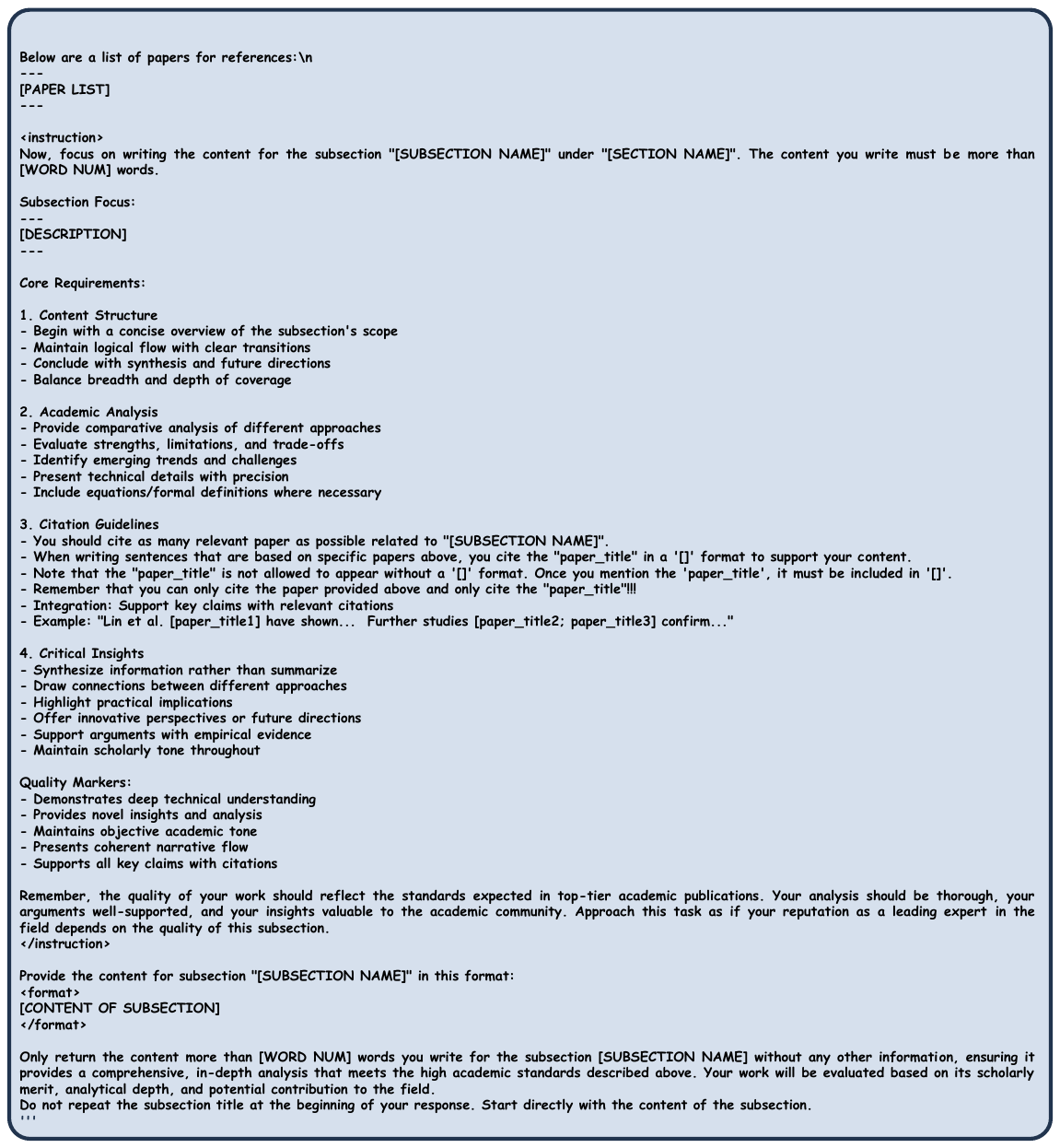}
\end{figure*}

\begin{figure*}[t!]
     \centering
     \includegraphics[width=0.98\linewidth]{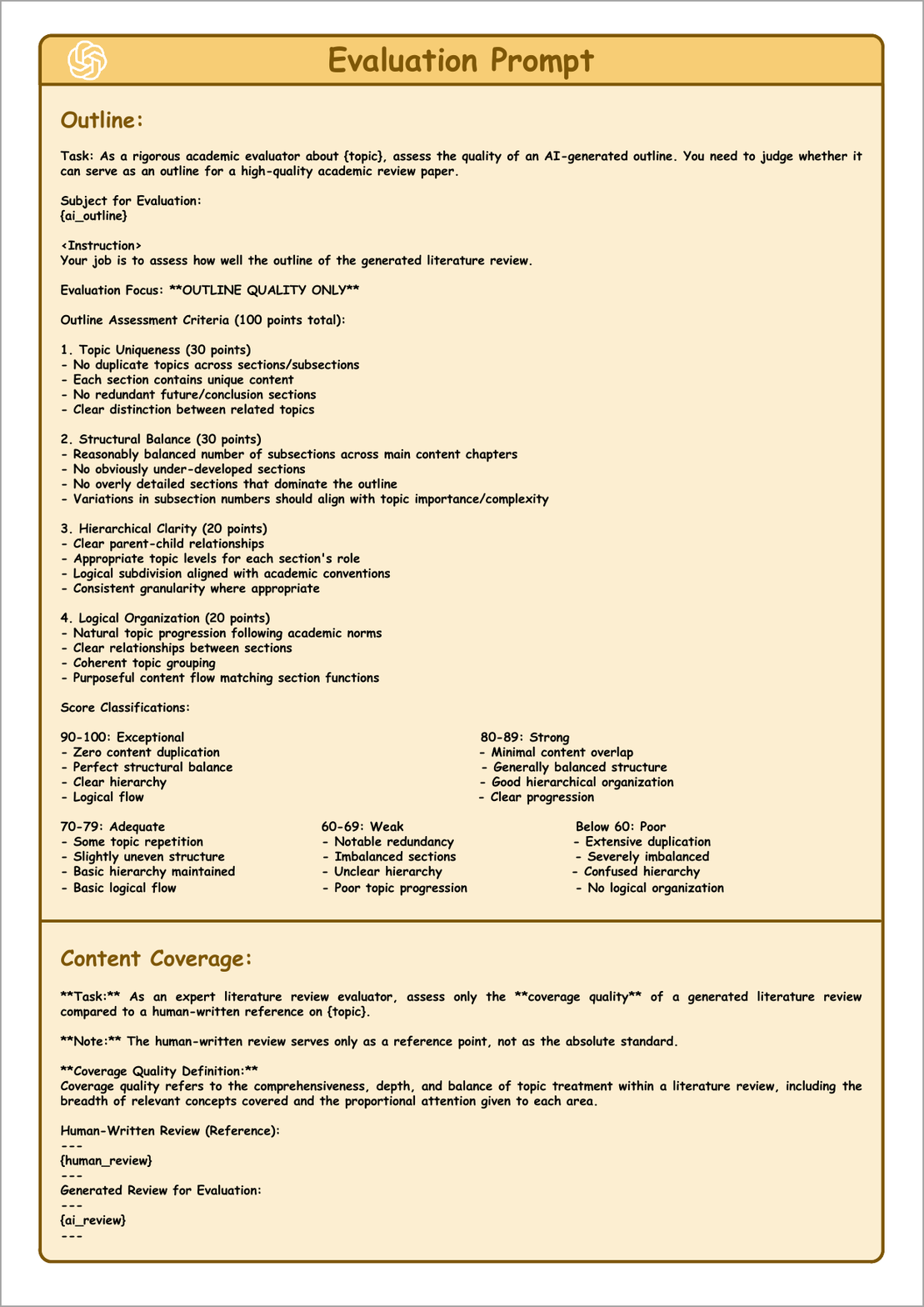}
\end{figure*}

\begin{figure*}[t!]
     \centering
     \includegraphics[width=0.98\linewidth]{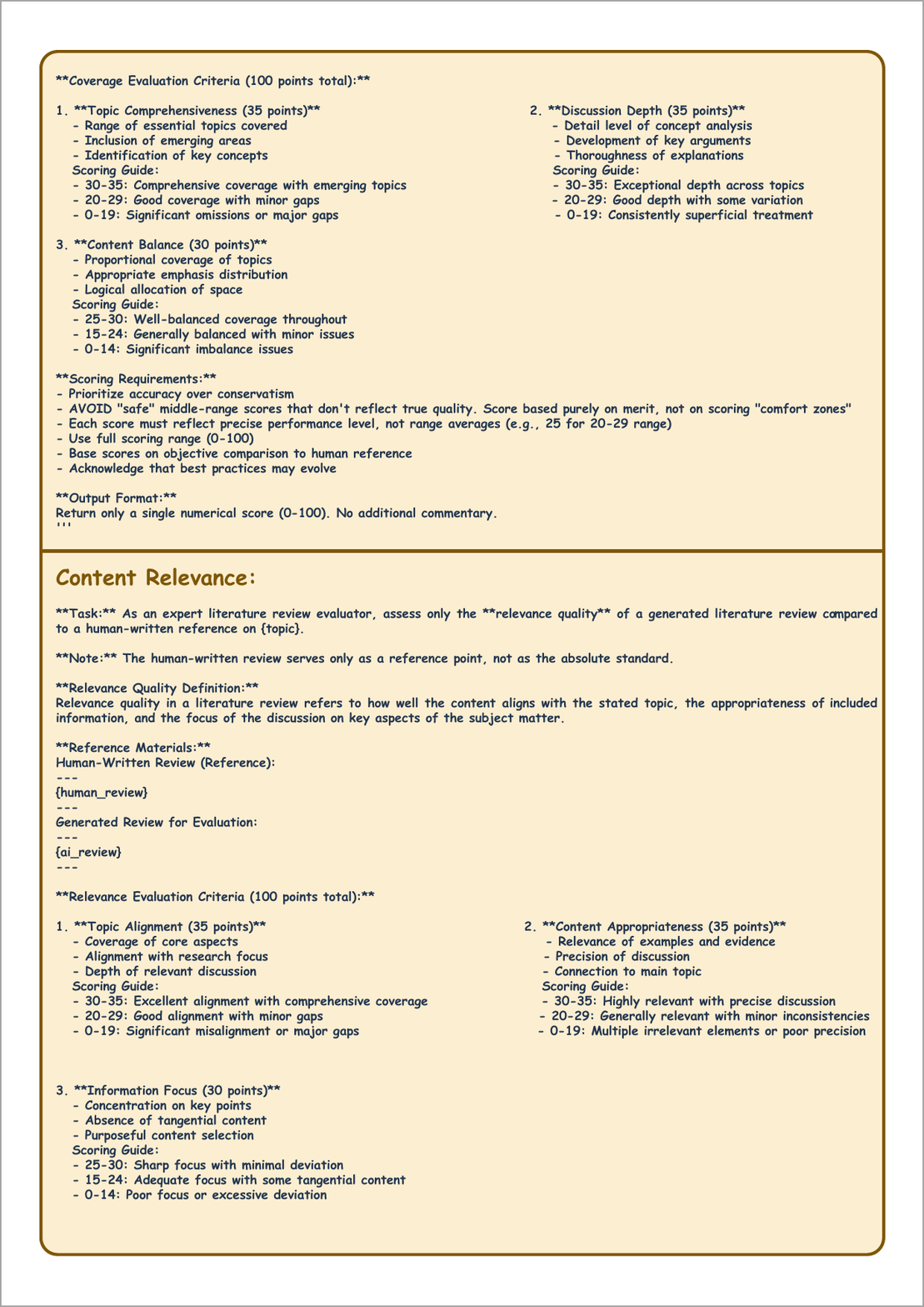}
\end{figure*}

\begin{figure*}[t!]
     \centering
     \includegraphics[width=0.99\linewidth]{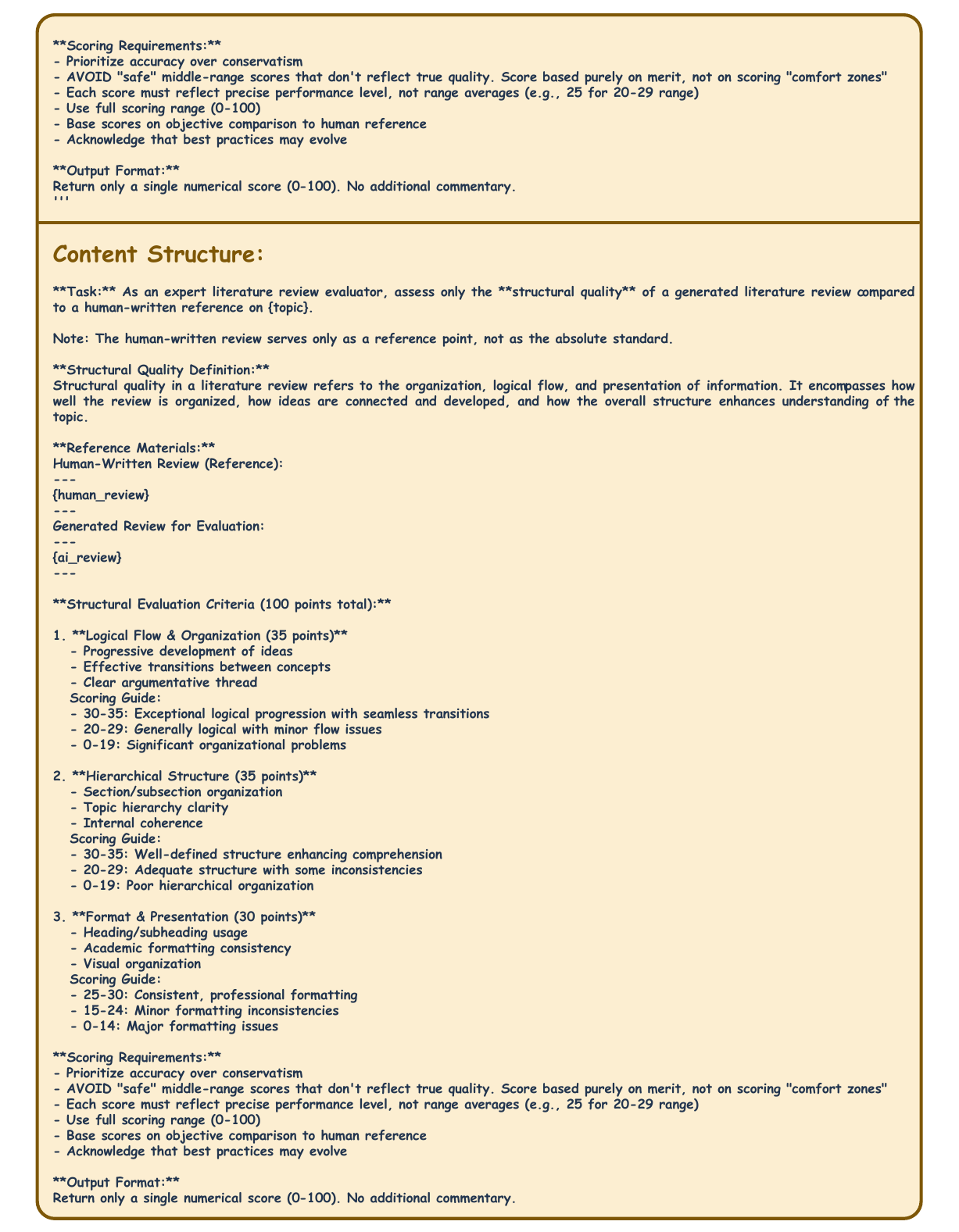}
\end{figure*}